\documentclass[a4paper,10pt,twocolumn]{article}
\usepackage[T1]{fontenc}
\usepackage[utf8]{inputenc}
\usepackage{times}
\usepackage[top=18mm,bottom=19mm,left=17.5mm,right=17.5mm]{geometry}
\usepackage{amsmath,amssymb}
\usepackage{graphicx,booktabs,tabularx,array}
\usepackage{microtype,caption,placeins,xurl,enumitem,ragged2e}
\usepackage{xcolor,titlesec,balance}
\usepackage[numbers,sort&compress]{natbib}
\usepackage[hidelinks]{hyperref}
\hypersetup{pdftitle={Collective Collaboration of Recursive Autonomous Agents},pdfauthor={Ziqing Zhou, Linqiang Hu, Yuning Chen, Hongda Zhang, Zhao Huang, Ding Zhou, Haining Xu}}
\titleformat{\section}{\large\bfseries}{\thesection}{.6em}{}
\titleformat{\subsection}{\normalsize\bfseries}{\thesubsection}{.6em}{}
\titlespacing*{\section}{0pt}{1.2ex plus .3ex minus .1ex}{.6ex plus .1ex}
\titlespacing*{\subsection}{0pt}{1ex plus .2ex minus .1ex}{.45ex plus .1ex}
\newcolumntype{Y}{>{\raggedright\arraybackslash}X}

\setlist{nosep,leftmargin=*}
\begin{document}
\twocolumn[{
\begin{center}
{\LARGE\bfseries Collective Collaboration of Recursive Autonomous Agents\par}
\vspace{.7em}
{\large
Ziqing Zhou\textsuperscript{$\dagger$},
Linqiang Hu, Yuning Chen, Hongda Zhang,\\[.25em]
Zhao Huang, Ding Zhou,
Haining Xu\textsuperscript{*}\par}
\end{center}
\vspace{.35em}
\begin{minipage}{\textwidth}
\textbf{Abstract.}
Smart factories must translate order-level objectives into production tasks at multiple scales and combine the capabilities of individual units into an overall delivery capability. To address repeated adaptation of organizational mechanisms across equipment, workshops, and factories, as well as coordination across levels under changing production conditions, this paper develops a framework for the collective collaboration of recursive autonomous agents (RAAs). A common task and capability interface supports the recursive decomposition of production objectives from the top down and the aggregation of conditional capabilities through peer collaboration from the bottom up. Composite units can therefore participate in higher-level organizations as agents of the same kind. We formulate organizational models for leaf and composite units, specify interface closure under composition and conditions for finite task expansion, and develop operational mechanisms linking task decomposition, resource commitments, result aggregation, and coordination within the scope affected by change. A production network integrating machining, inspection, and logistics across multiple workshops illustrates how a common organizational kernel is instantiated at the equipment-service, workshop, and factory scales. The framework connects recursive organization with explicit task responsibilities, capability conditions, and execution interfaces for manufacturing systems.
\par\vspace{.45em}
\end{minipage}
\vspace{1em}
}]
\begingroup
\renewcommand{\thefootnote}{\fnsymbol{footnote}}
\footnotetext[2]{%
First author: Ziqing Zhou.
Email:zhouziqing@fudan.edu.cn.}
\footnotetext[1]{%
Corresponding author: Haining Xu.
Email: xuhaining@jscltec.com.}
\endgroup

\section{Introduction}
\label{sec:intro}
Traditional factories typically organize production around process requirements and managerial responsibilities. Production management systems develop order and resource plans, workshops dispatch work to stations and equipment, and local control systems execute the required operations. This organization provides clear responsibilities and supports continuous production when products, processes, and resource configurations remain relatively stable. As batch sizes decrease, product mixes change, and delivery requirements become more demanding, production schedules must respond more frequently to equipment failures, fluctuations in processing rates, and material delays. Revising an existing plan therefore involves not only individual equipment operations but also interactions among process handovers, shared resources, and order due dates. Manufacturing control research has consequently expanded from plan execution within fixed structures to distributed decision-making and production-system reconfiguration in response to change~\cite{leitao2009survey,koren1999rms}.

Smart factories strengthen the connection between production operations and management through equipment connectivity, state sensing, cyber-physical systems, and data-driven decisions. The cyber-physical production systems discussed by Monostori et al. integrate computation, communication, and physical processes, providing a basis for synchronized production-state information and decisions at multiple scales~\cite{monostori2016cps}. The 5C architecture of Lee et al. organizes the progression from data connection and information conversion to cognition and configuration~\cite{lee2015cps}. These developments allow manufacturing resources to participate in task allocation according to their current states and extend production management from periodic planning to rolling optimization and event-driven rescheduling~\cite{vieira2003rescheduling}. Smart factories thus provide the technical foundations for connecting operational information with production decisions. Their further development requires a coherent organizational process that links local responses, collaboration among units, and overall delivery.

This coherence depends on how production units are organized as well as on decision quality. Local optimization by individual machines does not by itself establish a deliverable workshop capability. Similarly, information sharing among workshops still requires coordination of process handovers, transport windows, and shared resources. If equipment, workshops, and factories use separate task representations and coordination procedures, adding production units or changing management boundaries can require repeated adaptation of interfaces and organizational logic across levels. The resulting research question is how to reuse production-task organization as units are composed, while preserving differences in domain activities and communicating local changes to the appropriate coordination scope.

\begin{figure*}[t]
\centering
\includegraphics[width=\textwidth]{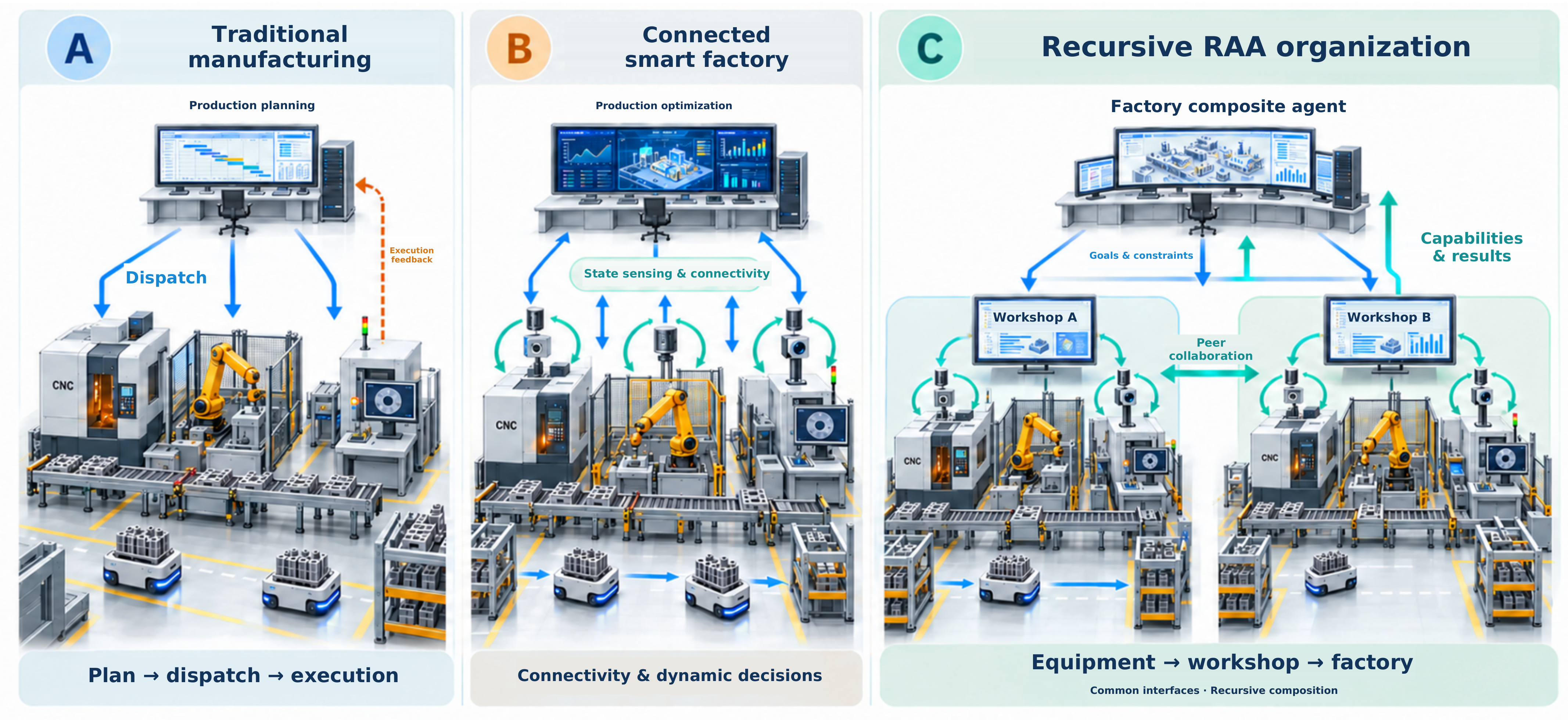}
\caption{Three representative forms of production organization. (a) Traditional manufacturing establishes vertical management through planning, dispatch, and equipment execution. (b) Smart factories use operational sensing, data connectivity, and decision services to support production optimization and dynamic adjustment. (c) Building on these technologies, RAA organization uses common task and capability interfaces across scales, enabling downward objective decomposition, peer collaboration, and upward aggregation of composite capabilities. The panels summarize organizational emphases; they neither imply that existing smart factories lack distributed coordination nor rank their performance.}
\label{fig:factory-overview}
\end{figure*}

Recursive autonomous agents (RAAs) organize production through individual autonomy and collective collaboration. Here, engineering autonomy denotes a unit's ability to select and revise its execution strategy within specified objectives, constraints, and authorization. Collective collaboration connects the tasks, resources, and handover conditions of multiple units so that their local execution jointly satisfies an overall objective. An RAA factory is therefore a recursively composed production organization: equipment or operational units collaborate with peers to establish workshop-level capabilities, and workshops further establish factory-level capabilities. The composite entity continues to participate in higher-level collaboration as an agent with a defined identity, capability, and responsibility boundary. Figure~\ref{fig:factory-overview} outlines the relationship among traditional manufacturing, connected smart factories, and recursive RAA organization.

The defining feature of recursion is that composition preserves the same kind of organizational capability. From the top down, order objectives are decomposed into workshop and equipment tasks, with subtasks retaining the same contract until they map to executable skills. From the bottom up, members establish joint arrangements subject to process, material, time, and resource conditions, after which a composite unit publishes its overall capabilities and execution results. Cross-level isomorphism supports this process through consistent semantics and lifecycles for objective acceptance, capability queries, task delegation, resource commitments, and result feedback. Domain knowledge, time scales, and solution algorithms may differ across levels; the reusable elements are organizational interfaces and operating rules.

Holonic and fractal manufacturing have already investigated the aggregation, self-similarity, and dynamic reconfiguration of autonomous entities. PROSA explicitly separates aggregation of manufacturing entities from control algorithms; fractal manufacturing builds self-similar organizations from recurring functional modules; and ADACOR examines the dynamic relationship between overall coordination and local self-organization~\cite{vanbrussel1998prosa,ryu2003fractal,leitao2006adacor,barbosa2015adacor}. Building on these foundations, this paper specifies RAA recursion through a common task contract, conditional capability aggregation, a reusable organizational kernel, and collaborative reconfiguration within the scope affected by change. The focus is on how recursion connects production-task expansion with the formation of composite capabilities.

The main contributions are: (1) a unified organizational model for leaf and composite agents, with conditions for cross-level interface consistency, closure under composition, and finite task expansion; (2) a recursive collaboration mechanism connecting task decomposition, peer negotiation, resource commitments, and result aggregation, including rules for determining coordination scope after task or capability changes; and (3) a cross-scale task mapping in a multi-workshop machining--inspection--logistics network, together with an implementation architecture based on a common organizational kernel and configurable domain components.

\section{Related Work}
\label{sec:related}
\subsection{Manufacturing Reconfiguration and Cyber-Physical Production Systems}
Adapting manufacturing systems to changing demand requires coordinated adjustments to physical resources, control functions, and production organization. Koren et al. introduced reconfigurable manufacturing systems, in which adaptable machines, controllers, and system-design methods allow production capacity and functionality to change with demand~\cite{koren1999rms}. This work connects reconfigurability with system structure: adaptation encompasses both runtime task rescheduling and changes in resource configurations and control structures. For collaboration across equipment and workshops, resource modularity provides a basis for composition, while capability descriptions, task delegation, and handover constraints must still be established among modules.

Cyber-physical production systems further connect manufacturing entities with their digital representations. Monostori et al. discuss advances in sensing, connectivity, and collaboration through the integration of manufacturing science, computation, and communication~\cite{monostori2016cps}. The 5C architecture of Lee et al. organizes data connection, conversion, networked analysis, cognition, and configuration to bring operational data into decision-making~\cite{lee2015cps}. Reconfiguration and cyber-physical integration thus provide complementary resource and information foundations for dynamic production organization. This paper examines how task and capability interfaces remain consistent as organizational scale changes and how composite units continue to participate in higher-level collaboration.

\subsection{Holonic Manufacturing and Fractal Organization}
Holonic manufacturing treats production entities as autonomous, cooperative units that are simultaneously wholes and parts of larger systems. PROSA, proposed by Van Brussel et al., comprises product, order, and resource holons, with staff holons providing specialized decision support~\cite{vanbrussel1998prosa}. Aggregation allows entities at different scales to retain distinct external identities, while specialization permits different manufacturing functions to use different internal implementations. PROSA also explicitly separates organizational structure from control algorithms, using self-similarity to support component integration and system reconfiguration. These principles are directly relevant to the encapsulation of composite units and reuse of mechanisms across levels considered here.

Fractal manufacturing likewise emphasizes self-similar functional organization. Ryu et al. describe basic units using five functional modules---observer, analyzer, organizer, resolver, and reporter---that handle production tasks through inter-agent collaboration and negotiation~\cite{ryu2003fractal}. Repeating a common functional structure across scales enables units to adjust internal behavior and interactions as conditions change. These studies establish a direct foundation for cross-scale self-similarity and functional reuse. This paper further describes their operational connections through task contracts, conditional composite capabilities, and commitment lifecycles.

ADACOR addresses the relationship between global coordination and local responsiveness. Its supervisor, task, product, and operational holons organize production and adjust coordination under normal and disturbed conditions to balance overall planning with local adaptation~\cite{leitao2006adacor}. The evolutionary architecture of Barbosa et al. extends self-organization to behavioral and structural adaptation, allowing both unit strategies and organizational relationships to change with production conditions~\cite{barbosa2015adacor}. Drawing on these ideas, the present work focuses on cross-level consistency in task decomposition, composite capabilities, resource commitments, and result feedback, and describes their operational integration through a common kernel.

\subsection{Hierarchical Task Planning and Capability-Driven Allocation}
Hierarchical task network (HTN) planning progressively expands compound tasks into primitive tasks using domain methods. SHOP2 selects decomposition methods according to the planning state and supports partially ordered tasks and their dependencies~\cite{nau2003shop2}. HTN provides a representation from production objectives to operation sequences, but task hierarchies and organizational hierarchies are distinct modeling constructs. A single planner may expand the entire task hierarchy centrally, or delegate tasks to other units for further planning. RAA adopts the latter organization: a composite assignee organizes its members under the same contract while retaining responsibility for fulfilling the parent task.

The Contract Net Protocol establishes distributed delegation through task announcements, bids, and awards; assignees can also organize further subcontracting~\cite{smith1980contract}. Recursive delegation is therefore not unique to RAA. Its application to manufacturing depends on a shared interpretation of task information, capability conditions, commitment updates, and handovers. The capability and skill reference model of K\"ocher et al. provides a common information basis for describing resource functions and callable implementations~\cite{kocher2023capabilities}. Building on this distinction, the present work separates available functions from the tasks a unit can currently accept, incorporating resource calendars, quality applicability, and handovers among members into composite capability descriptions.

Task structures can also constrain learning-based decisions. Mu et al. combine HTN with multi-agent reinforcement learning, using task decomposition and subgoals to organize cooperative policy learning~\cite{mu2025htnmarl}. Such approaches show how planning knowledge can connect to lower-level policies, while cross-level reuse of an organizational kernel requires a separate definition. RAA permits rules, HTN, mathematical optimization, and learned policies as internal solution methods, connected through common task, resource, and result contracts.

\subsection{Production--Logistics Coordination and Dynamic Scheduling}
Dynamic scheduling revises existing arrangements when production states change. Vieira et al. organize rescheduling research in terms of strategies, triggering policies, and methods, and discuss the relationship between schedule revision and stability~\cite{vieira2003rescheduling}. Event-driven scheduling, periodic rolling planning, and local schedule repair can be used within different organizational architectures. Dynamic responsiveness therefore depends jointly on scheduling policies, triggering mechanisms, and organizational structures. Architectural differences chiefly concern the distribution of information and decisions, which constraints are handled locally, and when coordination must be expanded.

Integrated production and logistics scheduling requires joint treatment of processing and transport resources. Gehlhoff and Fay investigate agent-based decentralized integrated scheduling, coordinating production and transport arrangements to address temporal and resource coupling in dynamic manufacturing~\cite{gehlhoff2020integrated}. Their work demonstrates that task allocation and logistics services must share feasibility conditions. In a recursive organization, aggregating processing capabilities without transport and handover conditions can produce an external capability description that exceeds what is executable. Capability aggregation must therefore retain external resource dependencies relevant to higher-level decisions.

Multi-agent reinforcement learning combines local policies with global guidance for large-scale scheduling. Jang et al. investigate agent coordination under directed acyclic graph (DAG) constraints, linking dependent subtasks through leader--follower roles and reward mechanisms~\cite{jang2025dag}. Their semiconductor-factory scheduling study further combines abstract global objectives with local equipment scheduling in a leader--follower structure~\cite{jang2025factory}. These approaches provide algorithmic context for hybrid production decisions, while scheduling performance and recursive reuse across levels remain distinct properties. This paper models organizational structure separately from local solvers, making coordination scope, capability encapsulation, and mechanism reuse explicit.

\subsection{Industrial Agent Implementation and Model-Assisted Planning}
Industrial agent deployment must connect operational interactions with heterogeneous automation equipment. Seitz et al. develop an automation-platform-independent multi-agent system, connecting production resources through common information representations and OPC UA communication, and evaluate it in order-driven production and adaptable-factory scenarios~\cite{seitz2021platform}. This work shows how common communication, resource self-description, and modular integration support heterogeneous production systems. Cross-level task responsibilities, composite capabilities, and lifecycles additionally require explicit organizational rules.

Large models offer new ways to connect unstructured production requirements with manufacturing capabilities. Lim et al. introduce large language models into multi-agent manufacturing, enabling product and resource agents to interpret user requirements and exchange task information~\cite{lim2024manufacturing}; their subsequent work investigates model-assisted adaptive task planning and coordination~\cite{lim2026adaptive}. PaLM-E jointly uses visual, state, and language information for embodied tasks, providing a methodological reference for incorporating multimodal operational information into planning~\cite{driess2023palme}. These approaches primarily extend task interpretation, knowledge use, and candidate generation. Organizational recursion concerns how task delegation and composite-unit identity remain consistent as scale increases.

Here, large models generate candidates within the organizational kernel, while explicit task models, process rules, and resource constraints determine execution conditions. Model-generated candidates undergo the same checks as those produced by rules or optimizers, and confirmed plans connect to equipment through skill interfaces. Semantic reasoning can therefore be configured according to task requirements, while recursive organization rests on common contracts, closure under composition, and lifecycle reuse across levels.

\section{Recursive Organizational Model of RAA}
\label{sec:definition}
\begin{figure*}[t]
\centering
\includegraphics[width=\textwidth]{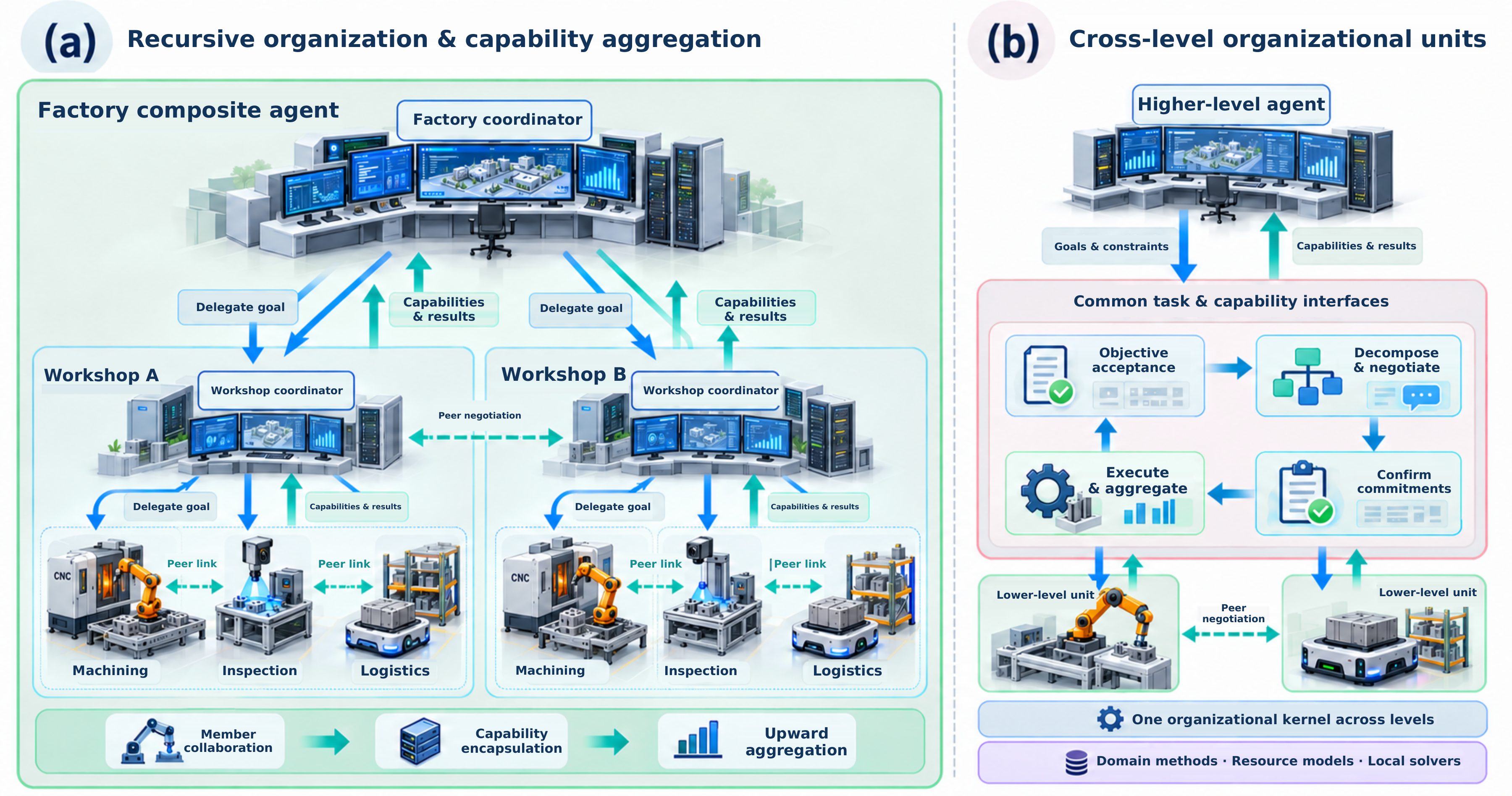}
\caption{Recursive RAA organization. (a) Operational units collaborate to form workshop and factory composite agents. Nested boundaries indicate organizational containment; blue arrows indicate objective delegation; teal arrows indicate capability and result feedback; dashed links indicate peer negotiation. (b) Each scale reuses task, capability, state, commitment, and result interfaces, together with an organizational kernel for objective acceptance, decomposition and negotiation, commitment confirmation, and execution and aggregation. Domain methods, resource models, and solvers are configured separately.}
\label{fig:recursive}
\end{figure*}

Figure~\ref{fig:recursive} presents the recursive organizational structure and instantiation of the common kernel. Equipment-service units form workshops, while workshops and shared services form a factory. Each composite unit maintains the same external interface and organizes collaboration among its internal members. Downward objective delegation and upward capability and result feedback connect scales; peer interactions coordinate task handovers and resources among members.

\subsection{Leaf Units, Composite Units, and Organizational Relationships}
A production-service unit with a defined identity, capability, responsibility boundary, and execution feedback is represented as agent $A_v$. Equipment participates as a leaf agent only when it supports a closed operational loop encompassing objective acceptance, state assessment, execution supervision, and result feedback. Sensors, actuators, and communication gateways participate through resource interfaces. The boundary of an autonomous unit is determined by an operational scope within which tasks can be independently accepted and their results verified. A leaf may correspond to a single machine or a group of devices required for a service, depending on that scope. Members form a composite agent under a coordinator:
\begin{equation}
A_v=\mathcal{C}\bigl(\{A_u:u\in\mathrm{ch}(v)\},\Theta_v\bigr),
\label{eq:composition}
\end{equation}
where $\mathrm{ch}(v)$ is the member set, $\Theta_v$ specifies membership, domain methods, handover conditions, resource-management responsibilities, and permissions, and $\mathcal{C}$ is the composition operation. The composite unit coordinates its members internally and accepts tasks, commits capabilities, and submits results externally under a single identity. The coordinator is an organizational role of the composite entity and does not provide additional physical capacity. Resource accounting must resolve to unique resource identifiers to avoid counting both a composite entity and its members as separate capacity.

Organizational containment is represented by a finite rooted tree $\mathcal{T}=(V,E_o)$ to make recursive expansion and responsibility assignment explicit. Peers interact through negotiation relations $E_h(t)$, while each task has a separate dependency graph $G_\tau$. The organizational tree, negotiation network, and task dependency graph specify coordination responsibilities, interactions, and process dependencies, respectively. A shared resource has one responsible manager and may serve different organizations; serving multiple tasks does not duplicate it into independent resources.

\subsection{Cross-Level Interface Consistency and the Organizational Kernel}
Cross-level isomorphism is defined here as consistency in interface semantics, organizational operations, and state transitions among units at different scales. It denotes a functional and protocol-level correspondence, allowing physical structures and workflow topologies to differ. Every unit uses the common interface schema
\begin{equation}
\begin{split}
\mathcal{I}=\langle &\mathrm{Goal},\mathrm{Capability},\mathrm{State},\\
&\mathrm{Commitment},\mathrm{Result}\rangle.
\end{split}
\label{eq:interface}
\end{equation}
These fields represent objectives and constraints, capabilities and applicability conditions, current state, active task and resource commitments, and completion status with acceptance evidence. Equipment, workshops, and factories may publish different service content, but the meanings, references, and versioning rules of the fields remain consistent. From a higher-level perspective, delegating an integrated machining-and-inspection service to a workshop follows the same kind of delegation, confirmation, and feedback rules as delegating a machining service to an equipment unit.

The common organizational kernel $\mathcal{K}$ handles objective reception, capability queries, candidate decomposition, negotiated scheduling, commitment confirmation, execution supervision, and result aggregation. Each unit instantiates the kernel using configuration $\theta_v$ and a local solver $\mathcal{S}_v$:
\begin{equation}
\Pi_v=\mathcal{K}(\mathcal{I};\theta_v,\mathcal{S}_v).
\label{eq:kernel}
\end{equation}
Cross-level reuse requires consistency in these organizational operations and their state-transition rules. Task decomposition methods, equipment models, queueing rules, transport planners, and optimization algorithms are configurable. Reusing a scheduling process means reusing problem submission, resource requests, commitment activation, and feedback updates; it does not require equipment-level path planning and factory-level scheduling to use the same numerical algorithm.

Interface consistency must be coupled with lifecycle reuse. Different solvers connect to the kernel through common task, resource, and result interfaces. The kernel maintains task states, delegation relationships, and commitment versions. Common interface names or message fields alone are insufficient: reuse also requires shared implementations of these operations and state transitions.

\subsection{Cross-Level Isomorphism in Workflow Networks}
\begin{figure*}[t]
\centering
\includegraphics[width=\textwidth]{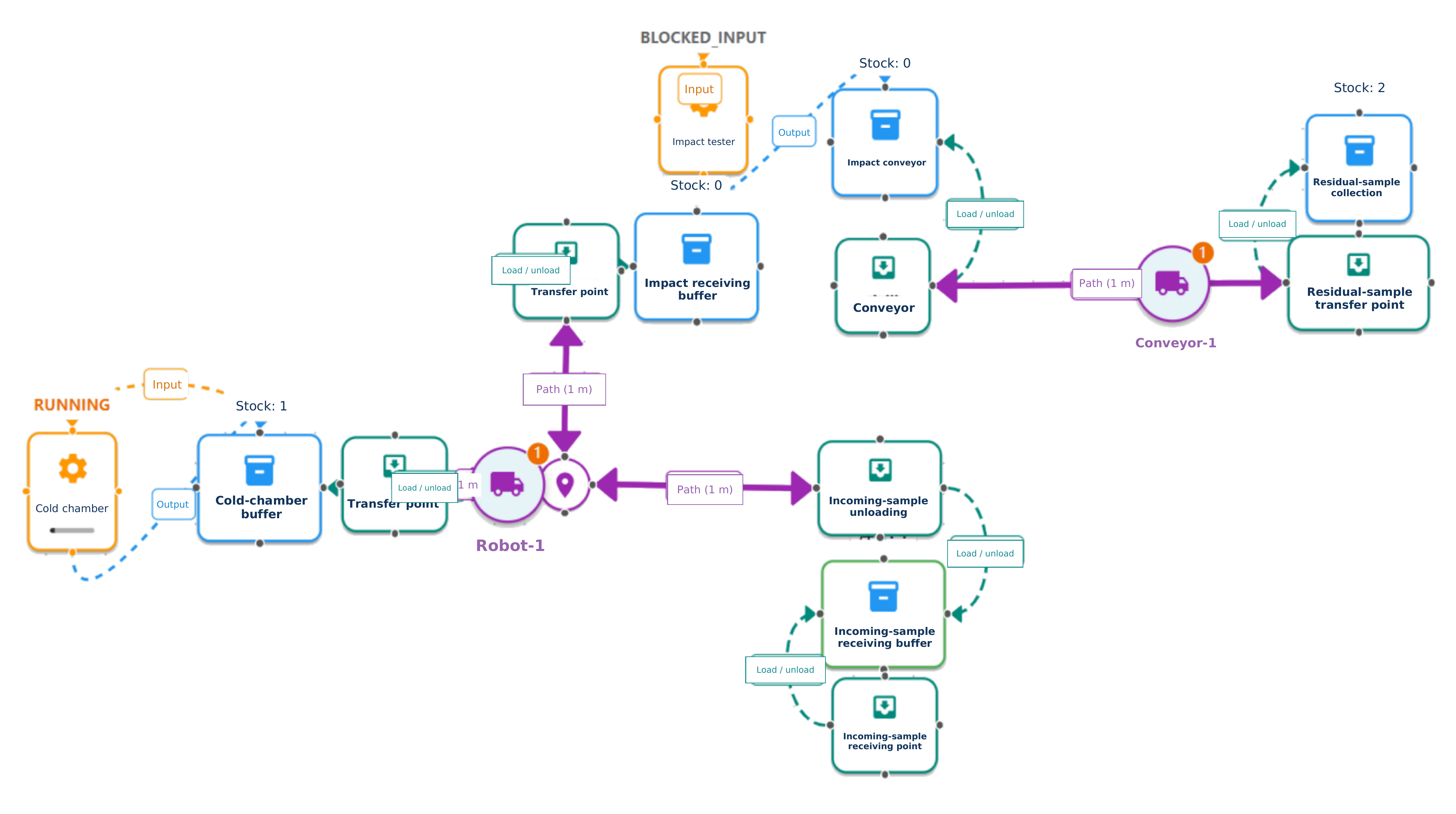}
\caption{Workflow network of a level-1 agent. Processing equipment, buffers, loading/unloading points, and transport resources form an operational unit. Orange and blue dashed links indicate input and output relationships in the original interface; teal dashed links indicate loading/unloading associations; purple links indicate transport paths. Inventory, operating states, and distances are snapshot values used solely to illustrate the operational structure.}
\label{fig:level1-workflow}
\end{figure*}

Workflow networks provide an intuitive illustration of cross-level isomorphism. Figure~\ref{fig:level1-workflow} shows the operational structure of a level-1 agent: low-temperature treatment, impact testing, buffers, loading/unloading points, robots, and conveyors jointly provide a service through material input, output, handover, and transport relationships. The level-1 agent is not identical to any single machine shown. It assumes overall responsibility for objective acceptance, resource organization, and result feedback within this collaborative scope. Externally, it provides a complete service capability; internally, it coordinates processing, storage, and transport resources.

\begin{figure*}[t]
\centering
\includegraphics[width=\textwidth]{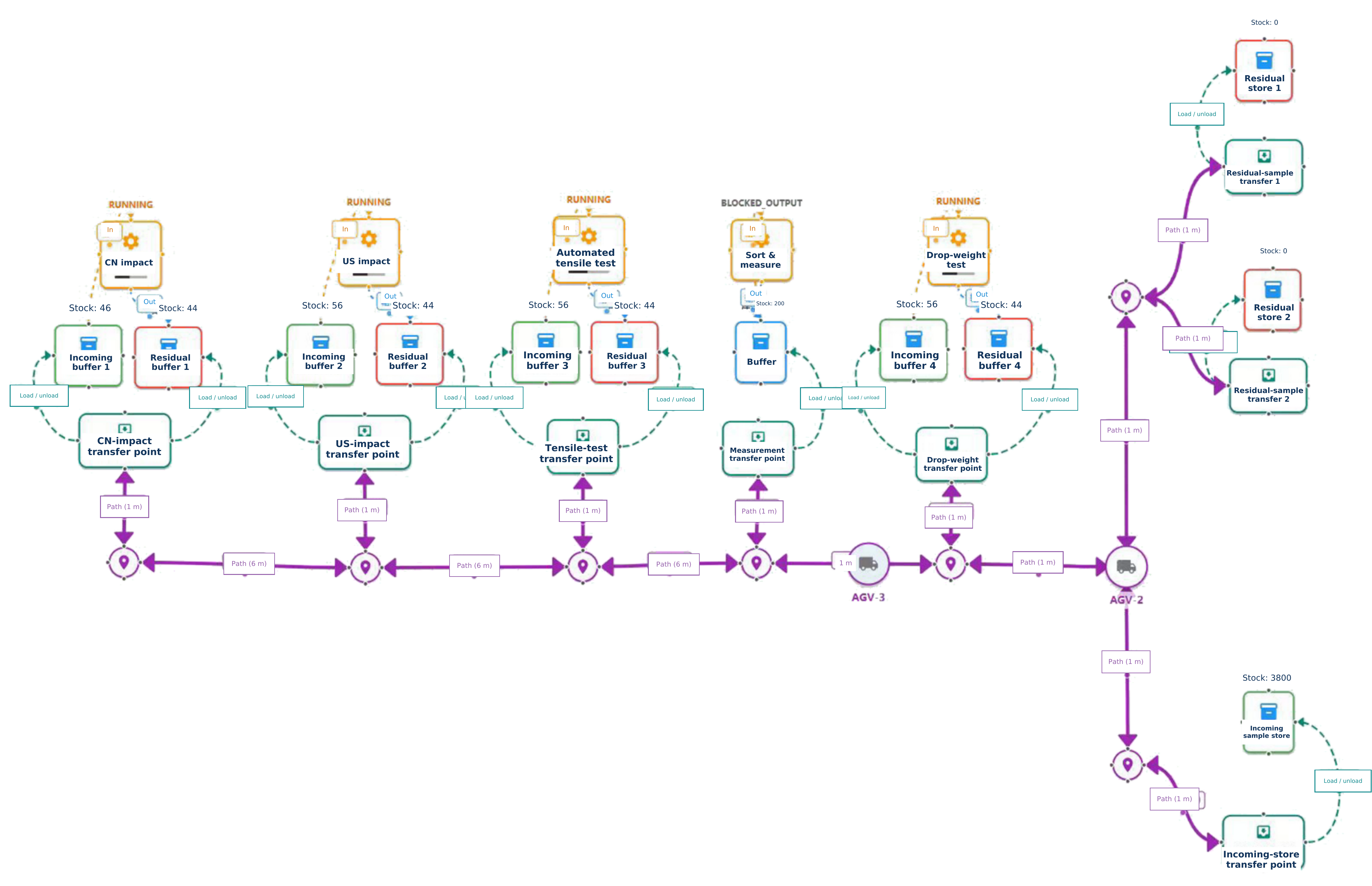}
\caption{Workflow network of a level-2 agent. Test and measurement service nodes form an integrated operating system through buffers, transfer points, and a shared AGV network. An encapsulated level-1 service can participate as a capability node at this level, reusing the roles and connection rules of input, processing, storage, handover, and transport at a larger scale. Internal implementations may differ among service nodes. Inventory, states, and path values are retained from the original interface snapshot.}
\label{fig:level2-workflow}
\end{figure*}

Figure~\ref{fig:level2-workflow} shows the operational structure of a level-2 agent. Service nodes for impact testing to Chinese and US standards, automated tensile testing, sorting and measurement, and drop-weight testing form a broader collaborative system with incoming-sample buffers, residual-sample collection, transfer points, and a shared automated guided vehicle (AGV) network. A service organized within the level-1 scope can be encapsulated as a processing node in the level-2 network. The level-2 agent again organizes input, service processing, buffering, handovers, and transport, and publishes an integrated service externally. Encapsulation requires registering the lower-level unit's capabilities and handover conditions with the corresponding service node. The different test nodes are not assumed to have identical internal equipment configurations.

The isomorphism illustrated by these networks lies in corresponding operational roles and organizational rules. Equipment-level processing can be aggregated into a service-level processing node; local buffering and handovers can be aggregated into inter-unit input/output conditions; and internal transport can be exposed as a queryable logistics service. Once the whole is encapsulated, the next level continues to organize tasks through the same roles of processing, storage, handover, and transport. The networks need not have the same number of nodes or the same topology. Purple paths and teal loading/unloading associations describe operational resource relationships rather than task delegation itself. Common contracts and the organizational kernel connect delegation to these resources and their service capabilities.

These role correspondences provide a structural basis for recursive encapsulation. A lower-level unit publishes its capabilities as a whole; a higher-level unit delegates a service through the common contract; and the assignee organizes execution through its internal workflow network before returning results to the parent task. Interface consistency, capability encapsulation, and the task lifecycle jointly turn this structural correspondence into an operational recursive process.

\subsection{Recursive Expansion, Upward Aggregation, and Termination}
Organizational recursion requires interface closure under composition: members satisfying the composition conditions form a whole through Eq.~\eqref{eq:composition} that continues to use Eq.~\eqref{eq:interface} in higher-level composition. Task recursion requires decomposed subtasks to retain the same task contract, allowing each assignee to choose between direct execution and further decomposition. The first requirement establishes the composability of composite units; the second connects production objectives to primitive operations.

Peer collaboration links these two directions. The presence of individual functions does not guarantee a deliverable composite capability. Machining and inspection equipment must also have compatible product scopes, material handovers, and transport conditions before jointly committing to a machining-and-inspection service. Members first establish executable internal arrangements through collaboration. The coordinator then abstracts these arrangements into an integrated service and reports its applicability, timing, and external resource dependencies. Cross-level isomorphism allows higher levels to organize this service in turn.

Recursion requires finite termination conditions. Organizational expansion can visit only members of the finite organizational tree. Within a unit, task methods must strictly decrease the rank of remaining compound tasks along every expansion branch until an approved skill with valid preconditions and permissions is matched. If no applicable method or leaf capability can be matched, the process reports infeasibility or insufficient conditions. Limits on decomposition depth, query time, and negotiation rounds bound computation; reaching a limit returns an incomplete status and its reason.

Equipment, workstations, lines, workshops, and factories are not mandatory fixed levels. Tightly coupled equipment may first form an operational module, with several modules forming a workshop, or a workshop may directly organize equipment with complete service capabilities. Levels depend on service responsibilities, resource coupling, and encapsulable capabilities. An additional composition level retains the existing interface and kernel while configuring the required domain knowledge, membership, and verification conditions.

\subsection{Structural Conditions for Recursive Organization}
Hierarchical organization assigns decision responsibilities at different scales; hybrid organization combines vertical coordination with local decisions; recursive organization additionally specifies interface closure of composite units and reuse of organizational processes across levels. RAA adopts a hybrid organization with recursive properties. Levels interact through common contracts and run the common kernel within their responsibility boundaries. Hierarchical depth can vary with operational scale, resource coupling, encapsulable capabilities, and fulfillment responsibilities.

Table~\ref{tab:recursive-properties} summarizes the structural and operational conditions. Common interfaces let members and composite units interact through the same kind of contract. Finite expansion and acceptance traceability connect tasks to execution outcomes. Preservation of these conditions after reconfiguration supports continued operation. These conditions characterize implementation consistency; production performance also depends on capability abstraction, resource coupling, and scheduling methods.

\begin{table*}[t]
\centering
\caption{Structural and operational conditions for recursive RAA organization.}
\label{tab:recursive-properties}
\small
\begin{tabularx}{\textwidth}{@{}p{2.8cm}YY@{}}
\toprule
Property & Required condition & Meaning in production organization \\
\midrule
Common task and capability interfaces & Leaf and composite units share contract and state semantics & Equipment, workshops, and factories can accept objectives, report capabilities, and submit results. \\
Closure under composition & A composite whole can participate as the same kind of member in higher-level collaboration & An added workshop or coordination level provides services under a single identity. \\
Reusable organizational operations & Objective acceptance, decomposition, negotiation, confirmation, and aggregation share a lifecycle & Domain algorithms connect as plugins without separate coordination procedures for every level. \\
Finite task expansion & Subtasks retain the same contract and expand to approved skills or return infeasibility & Objectives map to executable operations; unmatched capabilities are not silently bypassed. \\
Constraint and result traceability & Subtask results and handover evidence jointly map to parent-task acceptance conditions & Local completion supports an overall delivery decision. \\
Preservation after change & Revised membership and tasks still satisfy the interface and composition conditions & A reconfigured workshop participates in factory collaboration in the same way. \\
\bottomrule
\end{tabularx}
\end{table*}

\section{Capability-Driven Recursive Collaboration}
\label{sec:mechanism}
\begin{figure*}[t]
\centering
\includegraphics[width=\textwidth]{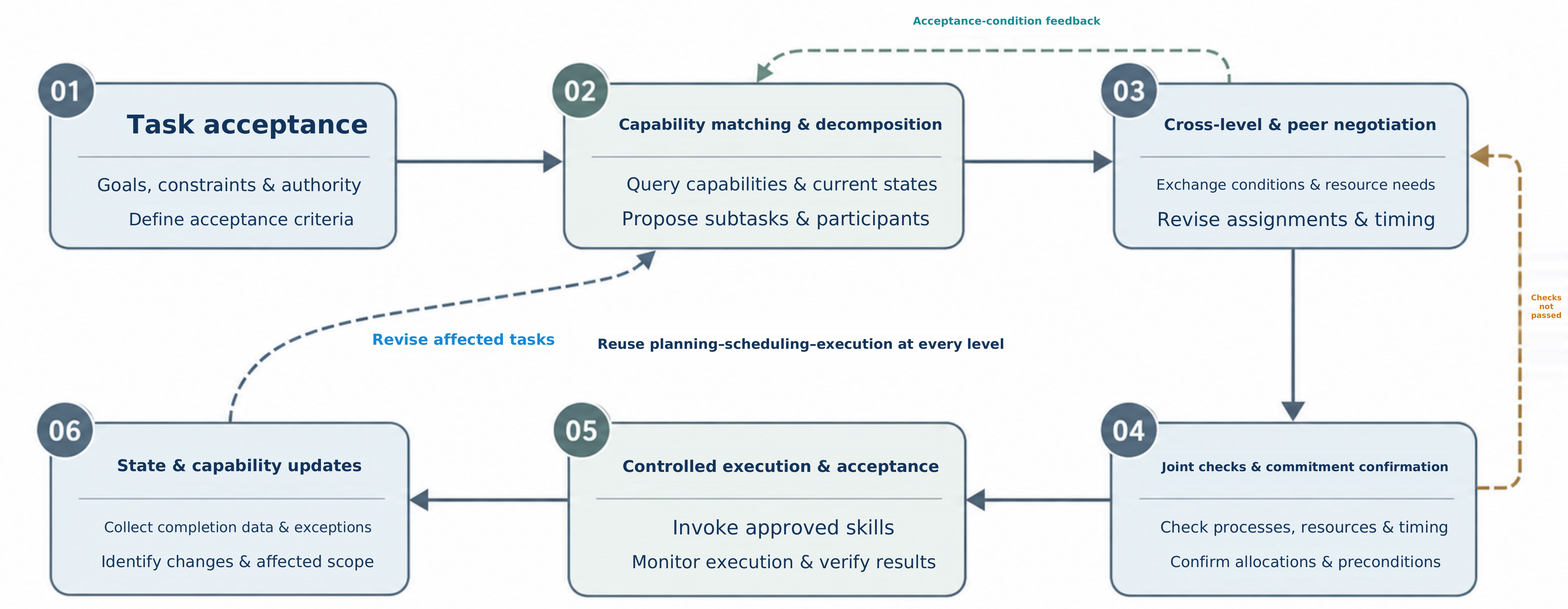}
\caption{Operational process of recursive collaboration. Task acceptance, capability matching and decomposition, negotiation, joint checks, controlled execution, and state updates form a closed loop at each level. Solid links indicate the main execution and feedback relationships; dashed links indicate condition feedback or plan revision after failed checks. Depending on the scope affected by change, the system uses internal adjustment, negotiation among affected peers, or higher-level coordination.}
\label{fig:workflow}
\end{figure*}

Figure~\ref{fig:workflow} presents the organizational lifecycle shared across scales: task acceptance, capability matching and decomposition, cross-level and peer negotiation, joint checks and commitment confirmation, execution and acceptance, and state updates. Equipment-service units, workshops, and factories handle different domain objects while retaining the same stage semantics and feedback relationships. The following subsections specify this lifecycle through task contracts, composite capabilities, resource constraints, and recursive calls.

\subsection{Common Task Contracts and Feasible Task Capabilities}
A task is represented as
\begin{equation}
\tau=\langle\mathrm{id},g,o,\Gamma,[r,d],q,b,\nu\rangle,
\label{eq:task}
\end{equation}
where $g$ is the objective; $o$ identifies a workpiece, batch, or material object; $\Gamma$ contains process, quality, resource, and safety constraints; $r$ and $d$ specify the release time and delivery requirement; $q$ gives acceptance conditions; $b$ specifies decision and execution authority; and $\nu$ is the version. Levels may refine objects and objectives, but tasks must retain references to their parents, applicable constraints, and result associations. Higher levels delegate required outcomes and their boundaries, while assignees select execution methods within the approved scope.

A unit's capabilities distinguish existing functions from the tasks it can currently accept. Let $\mathcal{F}_v(t)$ denote the set of task contracts for which $A_v$ can construct a feasible execution arrangement at time $t$, given its state, existing commitments, and specified resource conditions. This set is constrained by product scope, available skills, resource calendars, and processing rates. Implementations can use finite descriptions such as capability types, parameter ranges, available time windows, and conditional duration models rather than enumerating all tasks. After querying candidate capabilities, a higher-level unit uses commitment confirmation to turn conditional feasibility into a resource arrangement with a defined validity period.

Composite capability depends on feasible internal decomposition. Let $D_v$ produce subtasks and a dependency graph, $\sigma$ specify participants, timing, and resources, and $H_v$ express handover and cross-member constraints. The semantics are
\begin{equation}
\begin{aligned}
\tau\in\mathcal{F}_v(t)\ \Longleftrightarrow\ &\exists\,(D_v,\sigma),\quad \tau_u\in\mathcal{F}_u(t),\\
& H_v(D_v,\sigma)\ \land\ \mathrm{Cover}(D_v,\tau).
\end{aligned}
\label{eq:capability}
\end{equation}
Member feasibility must hold for every selected subtask assignee. $\mathrm{Cover}$ means that the decomposition outcomes and their handovers jointly cover the parent requirements. Member capability queries use the same candidate resource-allocation conditions, and the joint arrangement $\sigma$ rules out conflicting allocations. Equation~\eqref{eq:capability} specifies the semantic conditions for composite capability. An implementation publishes capability ranges supported by applicable methods or feasible arrangements and distinguishes an unresolved search from confirmed infeasibility.

\subsection{Top-Down Capability-Driven Task Decomposition}
After accepting a task, a unit checks its objects, objectives, and constraints, then queries member capabilities and existing commitments. Domain methods provide candidate decompositions, while current capabilities determine which methods and participants remain applicable. Planning produces a DAG $G_\tau=(V_\tau,E_\tau)$ whose edges describe process precedence or material handovers. Subtasks retain Eq.~\eqref{eq:task}. A composite assignee invokes the same kernel for further expansion, while a leaf selects approved skills, completing the mapping from task objectives to equipment execution.

Capability feedback can change the decomposition itself rather than only the assignee. Suppose machining and final inspection were originally assigned to one workshop. If its inspection resource becomes unavailable, the higher-level unit can reformulate the objective as machining, inter-area transport, and independent inspection instead of simply substituting a machine within the original task structure. This change is valid only if alternative inspection satisfies the original quality requirements, the workpiece can be transported, and transport resources are available. Domain methods and currently composable capabilities jointly determine task granularity, while mandatory quality steps and acceptance requirements remain in force.

Decomposition must also preserve applicable constraints. An overall due date can be allocated to local time windows where process relationships permit; quality requirements map to step-specific acceptance conditions; and authority can only be refined within higher-level authorization. The coordinator continues to maintain constraints that cannot be delegated independently to a single member, such as a shared resource limit across workshops. Each level records who checks a constraint, when it is confirmed, and which higher-level commitments would be affected by its violation, preventing global requirements from being lost during refinement.

\subsection{Peer Collaboration and Resource Commitments}
Peer units establish joint arrangements around handover conditions, execution windows, and shared resources. A machining unit reports expected completion and output state, a transport unit supplies service windows, and an inspection unit returns receiving and inspection requirements. Negotiation may use contract-net procedures, rules, or optimization, but every proposal must become a checkable set of tasks, time intervals, and resource requests. Different solvers enter execution through the same commitment-management process, allowing algorithms to be replaced while retaining the organizational mechanism.

For each process dependency $(i,j)\in E_\tau$, timing must satisfy at least
\begin{equation}
s_j\geq c_i+\ell_{ij},
\label{eq:precedence}
\end{equation}
where $s_j$ is the start time, $c_i$ the completion time, and $\ell_{ij}$ a handover or transport delay not modeled as a separate task. If transport is already an explicit task node, adjacent dependency constraints account for it to avoid double counting. For resource $k$, candidate arrangements must also satisfy
\begin{equation}
\sum_{i\in\mathcal{J}_k(t)}a_{ik}(t)\leq B_k(t),
\label{eq:capacity}
\end{equation}
where $\mathcal{J}_k(t)$ includes all tasks using resource $k$ under candidate and existing valid arrangements, $a_{ik}(t)$ is their resource usage, and $B_k(t)$ is available capacity. This scope includes requests from different parent tasks.

Once a proposal passes process, time, resource, and authority checks, its participants and responsible resource managers confirm a versioned commitment with a validity period. Members manage their internal resources; a unique manager confirms cross-member shared resources. Atomic calendar updates or equivalent mutual-exclusion mechanisms prevent concurrent requests from receiving conflicting allocations. Confirmation across multiple managers uses a commit process with timeout, cancellation, and recovery rules. The joint arrangement becomes effective only after all required parties confirm. The specific distributed commit protocol depends on implementation conditions.

Peers may confirm revisions within existing authorization and notify the coordinator if external commitments remain unchanged. Revisions affecting parent-task delivery conditions or resources committed to other tasks require a coordinator whose responsibility covers the affected parties. Thus, valid commitments and resource responsibilities determine the scope of local autonomy; a common organizational mechanism does not authorize any member to independently revise the overall plan.

\subsection{Bottom-Up Capability and Result Aggregation}
Capability aggregation translates internal collaboration into a service description usable by higher levels. Member capabilities, interface compatibility, resource conditions, and task methods jointly determine the objectives a composite unit can accept. The aggregate description retains at least service type, applicable products, quality scope, service windows, expected completion conditions, and external shared-resource dependencies. A composite unit may hide specific internal equipment choices, but it must expose shared constraints affecting other units' feasibility. Otherwise, higher levels may mistake conflicting local commitments for simultaneously executable arrangements.

Aggregation must retain the conditional nature of capability. Two machining units may each provide processing capacity, yet their joint service rate may be limited by a shared transport device. Likewise, idle machining and inspection equipment do not guarantee a feasible handover window before a given due date. Overall capacity and completion times depend on serial and parallel relationships, waiting, and shared-resource occupancy. When necessary, a composite unit returns a conditional quotation or feasible time window for a specific objective rather than a misleading scalar capacity.

Results return along task dependencies. Members submit evidence associated with objects, task versions, and acceptance conditions. The coordinator checks whether subtask results and handover records jointly satisfy the parent task. This relationship can be expressed as
\begin{equation}
\Bigl(\bigwedge_{i\in V_\tau}q_i(z_i)\Bigr)\land H_v^{\mathrm{exec}}
\ \Longrightarrow\ q_\tau(z_\tau),
\label{eq:verification}
\end{equation}
where $z_i$ is subtask result evidence and $H_v^{\mathrm{exec}}$ denotes handover consistency during execution. This implication is an acceptance relationship that the domain decomposition method must establish. Insufficient process conditions or evidence lead to a pending-verification status. Its validity depends on the specific method and measurement basis; no general formal guarantee for industrial quality requirements is asserted here.

Through successive capability and result aggregation, a factory can make delivery commitments based on integrated workshop services. If a workshop changes its internal equipment arrangement while retaining a valid external commitment, higher levels need not reconsider all internal details. Such encapsulation provides a basis for limiting reconfiguration scope, conditional on retaining the information needed in capability abstractions and on lower-level units being able to absorb changes within their commitments.

\subsection{Recursive Invocation of the Common Kernel}
These mechanisms form a task process $\mathcal{R}(A_v,\tau)$ applicable to any unit scale. If a task maps to an approved skill, the unit confirms resources and preconditions, executes it, and submits results. If member composition is required, the unit constructs subtasks and a joint arrangement, confirms commitments, invokes the same process on the subtasks, and aggregates their results. If neither an applicable skill nor a feasible decomposition is available, it reports unmet conditions:
\begin{equation}
\mathcal{R}(A_v,\tau)=
\begin{cases}
\mathcal{E}_v(\tau), & \text{skill executable},\\
\mathcal{V}_v\!\left(\{\mathcal{R}(A_u,\tau_u)\}\right), & \text{plan confirmed},\\
\mathrm{Fail}(\tau), & \text{not executable}.
\end{cases}
\label{eq:recursion}
\end{equation}
Here, $\mathcal{E}_v$ executes skills and checks results, while $\mathcal{V}_v$ aggregates results against parent-task acceptance conditions. Subcalls start according to task dependencies and resource arrangements; they need not run sequentially. Capability queries, decomposition, and negotiation before confirmation form the kernel's preparation process. Candidates without commitments do not enter physical execution. Tasks awaiting additional conditions remain pending. $\mathrm{Fail}$ indicates that the current invocation has not established an executable basis, rather than deleting the operational objective.

The same process in Eq.~\eqref{eq:recursion} applies to equipment services, workshops, and factories. A composite unit acts as a single invocation target when accepting a parent task and as an organizer when invoking its members internally. If an original condition becomes invalid during execution, the process suspends the affected branch and returns to preparation and confirmation at the appropriate scope. The members, subtasks, and versions used for result aggregation are updated accordingly. Recursion is thus realized as a shared invocation and state-transition process across scales.

\subsection{Recursive Reconfiguration Within the Affected Scope}
\begin{figure*}[t]
\centering
\includegraphics[width=\textwidth]{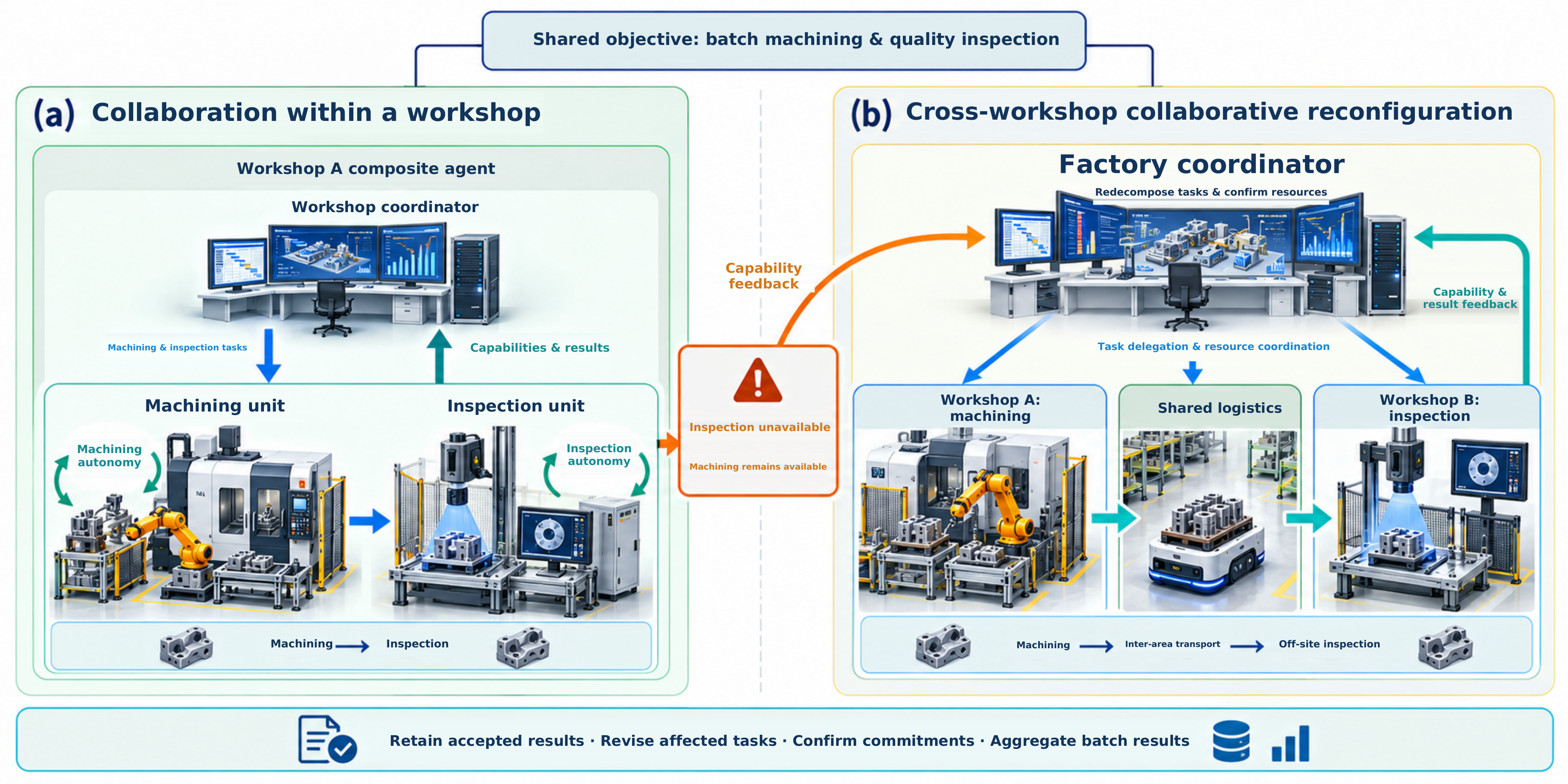}
\caption{Capability-driven collaborative reconfiguration across workshops. (a) Workshop A organizes machining and inspection internally and commits to batch delivery. (b) After its inspection resource is withdrawn, the affected objective is reported to the factory, which coordinates machining, shared logistics, and inspection in workshop B. Accepted results are retained; unexecuted tasks are reconfirmed under a new version; and overall quality and delivery requirements remain traceable. The figure illustrates the mechanism.}
\label{fig:reconfiguration}
\end{figure*}

Figure~\ref{fig:reconfiguration} illustrates how task structure and participant composition change together after inspection capability is withdrawn. Machining and inspection originally completed within one workshop are replaced by local machining, inter-area transport, and inspection in another workshop. The factory coordinator receives the constrained objective, decomposes the relevant tasks again, and confirms cross-workshop services. Each workshop continues to organize internal execution through the same kernel.

Production changes include updates to objectives and due dates, changes in equipment rates, resource withdrawal or addition, and invalid handover conditions. Event $e$ first updates directly associated state and capability descriptions. Its effects are then traced through task dependencies, shared-resource occupancy, and higher-level commitments, yielding the set of affected tasks and their units $\mathcal{A}_V(e)$. The analysis considers both dependencies and resource competition: changing one vehicle's occupancy may affect several workshops, while a local process delay may propagate through downstream tasks to order delivery.

If a unit can absorb the change internally while maintaining its external commitments, it invokes its kernel to revise internal arrangements. Changes involving peer handovers trigger negotiation among the relevant members. If the original composition no longer satisfies commitments or the necessary resources exceed local management authority, the unit escalates the constrained objective, alternative conditions, and confirmed progress. Higher levels receive this information through the same task and capability interfaces and repeat decomposition, negotiation, and confirmation, thereby implementing recursive reconfiguration during operation.

Given the organizational tree and resource responsibilities, an initial coordination scope can be determined by the lowest common ancestor of affected units and responsible resource managers:
\begin{equation}
v_e=\operatorname{LCA}\bigl(\mathcal{A}_V(e)\cup\mathcal{O}(e)\bigr),
\label{eq:scope}
\end{equation}
where $\mathcal{O}(e)$ is the set of relevant resource-management units. Equation~\eqref{eq:scope} determines an initial scope covering known dependencies and responsibilities. If higher-level commitments, authority, or newly identified coupling require broader coordination, the scope expands upward along organizational relationships. This rule assigns coordination responsibility; it is not a sufficient condition for globally optimal scheduling.

Reconfiguration proceeds from confirmed production states. Completed steps with valid acceptance results are retained; unstarted steps may be canceled or reassigned; and operations in progress are handled according to their interruptibility. Releasing old commitments and confirming new ones requires version linkage to prevent old instructions and revised plans from becoming effective simultaneously. Schedule stability also matters. Small changes recurring over a short interval can be consolidated within permitted tolerances and decision windows to reduce execution losses from frequent switching.

When members join or composition levels change, interfaces, capabilities, resource ownership, and task methods are verified before the common kernel organizes tasks. Equipment failure usually changes available capabilities and task arrangements without requiring a change to the organizational tree. Adding a workshop or changing management boundaries may alter organizational structure. These are distinct changes with different implications for scheduling and composition. If no feasible alternative exists, the unit reports waiting, delay, or an unattainable objective and updates the relevant delivery conditions.

\section{Production-Task Mapping}
\label{sec:scenario}
\subsection{A Multi-Workshop Machining--Inspection--Logistics Network}
An abstract production network comprising flexible workshops, shared transport units, and material buffers illustrates cross-level isomorphism and response to change. Orders contain batches with machining, inspection, and delivery requirements. Products may follow different process routes, and equivalent operations may run on multiple capable resources. Machining and inspection equipment have availability windows and processing rates. Transport resources are limited by load capacity, service duration, and concurrent occupancy. Tasks must jointly satisfy precedence, quality applicability, material handovers, resource exclusion, and due dates.

The scenario captures constraints common to multi-operation production, heterogeneous resources, and logistics across areas. Product routes, resource capabilities, and organizational configurations describe particular production instances. Logistics is represented through conditional service windows and transport times; path conflicts can be included as additional constraints. The network connects the recursive organizational model with concrete production tasks.

Equipment-service agents form leaf units. Machining, inspection, and internal transport units form workshop composite agents, while workshops and inter-area transport services are organized by a factory agent. Additional workstation or operational-module levels can be incorporated through the same composition operation. Internal and inter-area transport may have different managers or share one provider, provided that resource identities and allocation responsibilities remain unique and joint calendars prevent double counting.

\subsection{Cross-Scale Order Expansion and Capability Aggregation}
Consider the root task of machining and inspecting a product batch and delivering it to the outbound area before its due date. The factory queries workshop capabilities and inter-area transport conditions to construct batch allocations and delivery arrangements. A workshop receives a contract specifying products, quantities, quality requirements, and deadlines. It then queries internal equipment and logistics capabilities and decomposes the task into machining, handling, inspection, and related subtasks. Equipment units map their subtasks to approved operations executed through local control systems.

Peer collaboration occurs in the same way at each scale. Equipment units coordinate machining outputs and transport reception; workshops coordinate inter-workshop process handovers; and the factory coordinates integrated workshop services and outbound transport. Although domain content differs, each follows objective reception, capability feedback, candidate arrangement, resource confirmation, and result return. Table~\ref{tab:production-mapping} shows the same organizational mechanism instantiated across scales.

\begin{table*}[t]
\centering
\caption{Instantiation of the common recursive mechanism across production scales.}
\label{tab:production-mapping}
\small
\begin{tabularx}{\textwidth}{@{}p{2.3cm}YYY@{}}
\toprule
Operation & Equipment-service unit & Workshop composite unit & Factory composite unit \\
\midrule
Objective acceptance & Machine or inspect a specified workpiece & Complete batch machining and inspection & Fulfill orders and outbound delivery by due dates \\
Capability query & Skills, tools, materials, and availability windows & Equipment, internal logistics, and process feasibility & Workshop services, inter-area logistics, and batch acceptance conditions \\
Task decomposition & Approved operations and pre-/postconditions & Machining, handling, inspection, and dependencies & Batches, cross-workshop services, and delivery transport \\
Peer collaboration & Material and resource handovers among equipment units & Process and logistics windows among workshops & Composite services and shared resources within factory scope \\
Commitment confirmation & Local resources and execution permission & Member acceptance and internal resource calendars & Workshop commitments and shared inter-area resources \\
Upward aggregation & Operation results and current service capability & Batch results and conditional workshop capabilities & Order results and overall delivery capability \\
\bottomrule
\end{tabularx}
\end{table*}

In upward aggregation, equipment publishes current functions and service conditions. Workshops combine process methods and internal logistics into batch-level capabilities, and the factory combines workshops and delivery logistics into order-level capabilities. A higher-level unit treats a workshop as one assignee without reinterpreting every equipment operation at each decision. Abstractions retain external dependencies on cross-workshop shared resources so that encapsulation remains consistent with resource allocation.

\subsection{Changing Production Conditions and Organizational Adjustment}
The first type of change is a task update. New orders, revised quantities, and changed due dates alter root or batch contracts. If a workshop can absorb the change within its existing commitments, it updates only internal subtasks. Changes affecting other batches, inter-area transport, or factory delivery require reassignment by the appropriate higher-level unit. Frequent updates require a balance between continual revision and execution stability.

The second type is a reduction in processing rate. When processing time increases, the equipment-service unit first revises its expected completion conditions. A workshop may reassign work to an internal alternative while maintaining external commitments. If it can no longer meet the batch due date, it reports the capability change to the factory. The factory checks whether other workshops and inter-area transport provide a feasible alternative. Equipment, workshops, and the factory invoke the same organizational operations within their respective scopes, with the validity of external commitments determining how far coordination extends.

The third type is a change in resource conditions. After an inspection device is withdrawn, an internal machining-and-inspection task may be replaced, where feasible, by local machining, inter-area transport, and external inspection. Additional equipment or transport resources can expand the available compositions. Such changes may alter both the task structure and the resulting composite capability.

The fourth type is organizational expansion. An existing equipment network may be composed into a new workshop, or an intermediate coordination module may be added among workshops. The resulting composite unit retains the common interface and kernel. Expansion still requires capability registration, membership configuration, and verification; recursive composition does not remove these domain and integration responsibilities.

\section{Implementation Architecture}
\label{sec:implementation}
\subsection{Common Organizational Kernel and Domain Plugins}
At the implementation level, Eq.~\eqref{eq:kernel} maps to a common runtime kernel for task-object management, capability registration, membership queries, negotiation messages, commitment versions, event routing, and result aggregation. Equipment and workshop units use the same object and lifecycle implementation. Domain plugins supply machining methods, inspection rules, transport services, resource models, and applicable constraints. Changes in product mix are represented through applicable methods and member configurations; new processes extend the corresponding methods, skills, and verification basis.

Scheduling is separated from organizational management. The kernel provides solvers with task dependencies, candidate capabilities, resource calendars, and objectives, and receives feasible arrangements, unmet conditions, or infeasibility explanations. Rules, mathematical optimization, and learned policies can all provide candidates through this interface. Algorithm adaptation may involve interface conversion, parameter adjustment, domain modeling, and model retraining. Kernel reuse applies primarily to interfaces and coordination processes; a policy's suitability for a new scale or resource distribution remains a separate concern.

Composite units publish external capabilities and dependencies through a capability registry, with internal state updated by events. Capability abstraction must match task needs. An overly coarse abstraction can omit shared-resource constraints, whereas an overly detailed one exposes internal complexity to higher levels. Objective-specific queries can return conditional acceptance proposals after a higher-level unit states its requirements, combining capability encapsulation with necessary information exchange.

\subsection{Multimodal-Assisted Planning and Local Execution}
\begin{figure*}[t]
\centering
\includegraphics[width=.98\textwidth]{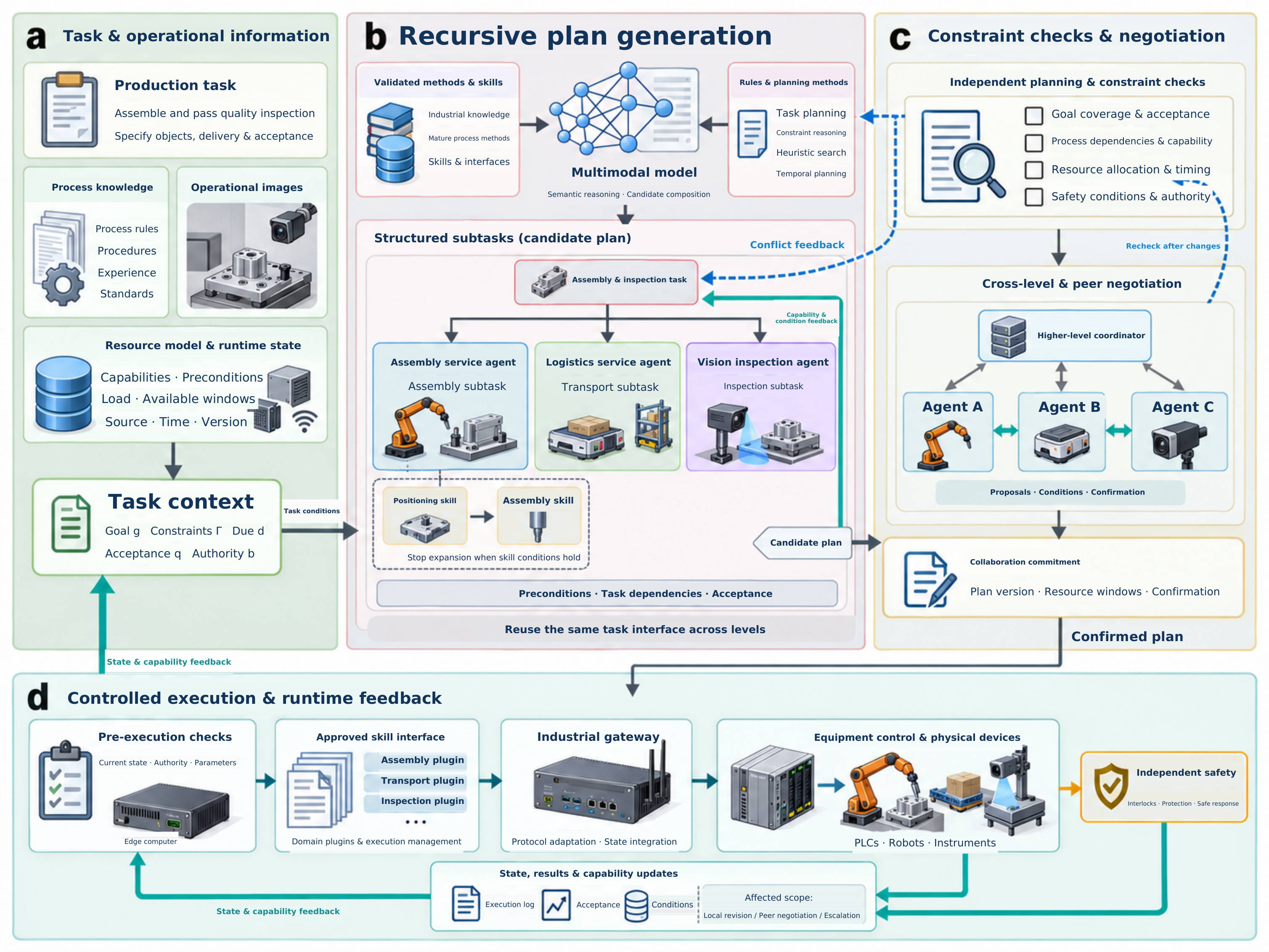}
\caption{Multimodal-assisted implementation of recursive collaboration. (a) Tasks, process knowledge, operational images, and resource states form a common context. (b) Models and planning methods generate structured subtasks that expand recursively where needed. (c) Capability, process, resource, and authority checks, together with cross-level and peer negotiation, establish commitments. (d) Confirmed tasks enter local execution through skill interfaces and industrial gateways, with state, capability, and acceptance results returned to the kernel. Assembly illustrates one domain plugin; validated service agents represent verified operational units. Edge computers and gateways are replaceable computing and integration components. This is an implementation schematic, not a claim that every module shown has been deployed.}
\label{fig:multimodal}
\end{figure*}

Figure~\ref{fig:multimodal} presents a multimodal-assisted implementation of recursive collaboration. Task descriptions, process knowledge, operational images, and resource states are first associated with a common task context. A multimodal model combines with rules or planning methods to generate candidate decompositions. Candidates become structured tasks with objectives, constraints, participants, and dependencies, then proceed through capability matching, resource checks, and negotiated confirmation. Semantic processing connects unstructured inputs to the organizational kernel, and specific models can be replaced according to operational needs~\cite{lim2024manufacturing,lim2026adaptive}.

Rules, HTN, or optimizers can directly handle structured orders and known processes. Multimodal models are useful where text, images, or heterogeneous knowledge require interpretation. Their scope is configured independently of organizational recursion. All candidates pass through common task objects and constraint checks before commitment management, so different generation methods do not change task responsibility or acceptance rules. Cross-level reuse resides in the organizational kernel, while semantic models provide revisable candidates.

Local computation can use dedicated edge computers or existing industrial computing platforms. Domain plugins hold methods and models; industrial gateways map PLC, robot, instrument, and logistics interfaces; and existing controllers execute approved skills. Before execution, the system checks task versions, resource permissions, and physical preconditions. Operational coordination, equipment control, and independent safety functions operate at their respective time scales. Research on learning-based robotic control likewise emphasizes coordination between learning mechanisms, constraints, and safety-control layers~\cite{brunke2022safe}. The implementation follows this separation of responsibilities: higher-level semantic reasoning does not replace local control or safety functions.

\subsection{Version Management and Execution Consistency}
Tasks, capabilities, domain methods, and organizational configurations retain version links. Resource changes can update current availability, while modifications to skills or capability boundaries require verification before publication. Composite capability versions record the member and method versions on which they depend, allowing higher levels to determine whether existing commitments remain valid. State updates that do not affect current tasks need not immediately trigger global revision. Changes to commitment conditions are handled using the scope rules in Section~\ref{sec:mechanism}.

After communication timeouts or node recovery, workpiece states, execution receipts, task versions, and resource occupancy are reconciled before continuation or cancellation. Requests carry unique invocation identifiers so that message retries do not trigger duplicate physical operations. If a coordinator fails, the scope in which members can continue is determined by valid commitments and locally verifiable conditions. New cross-unit commitments require restored coordination or an explicit takeover satisfying the specified takeover and consistency conditions.

Deployment verification covers at least interface semantics, termination of task expansion, constraint inheritance, unique resource identities, and result traceability. It also checks resource failures, task updates, and version transitions after reconfiguration. Correct capability composition must be verified against domain methods. A common kernel permits reuse of verification tools and some test scenarios, while new processes, physical interfaces, and operational boundaries require their own verification.

\section{Discussion}
\label{sec:discussion}
\subsection{Organizational Reuse and Engineering Adaptation}
Recursive design extracts objective acceptance, task delegation, commitment confirmation, and result aggregation from scale-specific implementations into a common organizational process. When a composition level is added, the composite unit retains existing contracts and represents its internal structure through membership and domain configurations. This reuse provides a basis for reducing cross-level interface adaptation. Its practical value depends on the operational scope covered by the kernel and the modularity of existing systems.

Domain processes, physical interfaces, and operational constraints still require modeling and verification. New products may require additional decomposition methods, new equipment needs state and skill mappings, and learning policies may require retraining after resource distributions change. Interface adaptation, domain modeling, equipment integration, and algorithm adjustment therefore remain distinct engineering responsibilities. Additional levels also introduce query, negotiation, and confirmation overhead.

\subsection{Dynamic Responsiveness and Schedule Quality}
When external commitments remain valid, local reconfiguration can preserve other units' arrangements and limit change propagation. Restricted coordination may, however, miss better solutions requiring resource exchanges across units. Capability aggregation reduces details exposed to higher-level solvers but can introduce abstraction error. Faster response therefore does not by itself establish lower tardiness, higher throughput, better resource utilization, or more stable schedules.

Coordination within the affected scope is most applicable when dependency information is sufficient, resource responsibilities are explicit, and disturbances can be absorbed locally. Strong coupling through shared resources or simultaneous invalidation of multiple delivery commitments requires broader coordination at a higher level. For small problems with centralized information, centralized solution methods may incur less runtime overhead. Frozen windows, adjustment thresholds, and decision deadlines can suppress repeated rescheduling; their effects are distinct from the recursive structure itself.

\subsection{Applicability and Limitations}
The model uses a finite rooted tree for primary containment relationships and a unique responsible manager for resources shared across organizations. This representation simplifies task responsibility and coordination-scope determination. Resource coalitions belonging simultaneously to several organizations require extensions to containment, authority inheritance, and commitment management. Capability composition also depends on domain methods and handover conditions. Unknown processes, measurement errors, and unmodeled coupling can affect feasibility judgments.

PROSA, fractal manufacturing, and ADACOR provide direct foundations for the framework. RAA is specified here through the connection among common contracts, conditional composite capabilities, and a common organizational kernel. The present contribution comprises an organizational model, operational mechanisms, production-task mappings, and an implementation architecture. These establish how recursive collaboration is organized without establishing quantitative advantages in development effort or dynamic scheduling performance.

\section{Conclusion}
\label{sec:conclusion}
This paper develops a recursive organizational model and operational mechanisms for collective RAA collaboration across production scales in smart factories. Common contracts expand production objectives into executable tasks from the top down. Peer collaboration establishes conditional capabilities, and composite units participate in higher-level organization as agents of the same kind. Consistent cross-level interfaces and lifecycles connect task decomposition, resource commitments, execution supervision, and result aggregation, allowing equipment-service units, workshops, and factories to reuse a common organizational kernel.

The framework specifies conditions for composite capabilities, finite expansion, shared-resource allocation, and coordination within the scope affected by change. A multi-workshop machining--inspection--logistics network illustrates the corresponding production-task mapping. Multimodal-assisted planning, domain solvers, and local execution interfaces provide configurable implementation support. Together, these elements connect recursive composition with explicit operational responsibilities and execution conditions in manufacturing organizations.

\FloatBarrier
\balance
\bibliographystyle{unsrtnat}
\bibliography{references}
\end{document}